\documentclass[letterpaper, 10 pt, conference]{ieeeconf}  
\usepackage{graphicx}
\usepackage{multicol}
\usepackage{amssymb}
\usepackage{bbm}
\usepackage{amsmath}
\usepackage{xcolor}
\usepackage{booktabs}
\usepackage{mathtools}
\usepackage{subcaption}
\usepackage{float}
\usepackage[ruled,vlined,linesnumbered]{algorithm2e}
\usepackage{hyperref}
\usepackage[english]{babel}
\newtheorem{theorem}{Theorem}
\usepackage{ulem}

\newcommand{\q}{\textbf{q}}

\IEEEoverridecommandlockouts                              

\title{\LARGE \bf

Learning State Conditioned Linear Mappings for Low-Dimensional Control of Robotic Manipulators

}

\author{Michael Przystupa$^{*\dagger}$, Kerrick Johnstonbaugh$^{*\dagger}$, Zichen Zhang$^{\dagger}$, Laura Petrich$^{\dagger}$, \\ Masood Dehghan$^{\dagger}$, Faezeh Haghverd$^{\dagger}$,
Martin Jagersand$^{\dagger}$ 
\thanks{$^{\dagger}$ Department of Computing Science,
        University of Alberta, Edmonton AB., Canada, T6G 2E8.{ 
           \tt\small \{przystup, kerrick, zichen2, llorrain, masood1, haghverd, mj7, \}@ualberta.ca
        }
       }%
\thanks{$^{*}$ Equal contribution} 
}

\begin{document}

\maketitle
\thispagestyle{empty}
\pagestyle{empty}


\begin{abstract}
Identifying an appropriate task space that simplifies control solutions is important for solving robotic manipulation problems. One approach to this problem is learning an appropriate low-dimensional action space. Linear and nonlinear action mapping methods have trade-offs between simplicity on the one hand and the ability to express motor commands outside of a single low-dimensional subspace on the other. We propose that learning local linear action representations that adapt based on the current configuration of the robot achieves both of these benefits. Our state-conditioned linear maps ensure that for any given state, the high-dimensional robotic actuations are linear in the low-dimensional action. As the robot state evolves, so do the action mappings, ensuring the ability to represent motions that are immediately necessary. These local linear representations guarantee desirable theoretical properties by design, and we validate these findings empirically through two user studies. Results suggest state-conditioned linear maps outperform conditional autoencoder and PCA baselines on a pick-and-place task and perform comparably to mode switching in a more complex pouring task.
\end{abstract}


\section{Introduction} \label{sec:introduction}

Robotic manipulation tasks involve controlling many actuators, or degrees-of-freedom (DOFs), simultaneously to perform complex motions like pouring liquids \cite{pouring2019yongqiang}, unscrewing light bulbs \cite{manschitz2014lightbulb}, or even playing table-tennis \cite{mulling2013tabletennis}. 
These complex tasks are achievable by various methods from domains like control theory, visual servoing, reinforcement learning, or teleoperation. For control solutions that involve learning agents (artificial or human), the consequences of poor design decisions can have adverse effects in robotic manipulation problems, where safety \cite{tosatto2021contextlampo} and sample efficiency \cite{li2021breakingsamplecomplexity} are desirable properties. In reinforcement learning, reducing sample complexity in environments with high-dimensional action spaces is a fundamental challenge \cite{chandak2019learningactionrepr}. For humans, previous research suggests that common control approaches such as mode-switching can be cognitively strenuous, with basic tasks like opening a door or picking an object requiring 30-60 mode switches \cite{herlant2016timeoptimalswitching,routhier2010usability}. 

\textit{Action maps} that transform low DOF inputs directly to relevant high-dimensional robotic commands
have been explored both to reduce cognitive strain during teleoperation and to increase the learning efficiency of artificial agents \cite{colome2015dimreducPMP,nutan2016dynamicMP,losey2021learnlatentLONGpaper}. The assumption is that joints work together in the high-dimensional ambient space to produce concerted motions that lie on some lower-dimensional manifold. Previous research on hand poses -- a high-DOF setting -- supports this assumption,  finding that principal component analysis (PCA) \cite{pearson1901fittingplanesPCA,Hotelling1933AnalysisOA} can capture 80\% of configurations with only two principal components \cite{santello1998posturalhands}. Authors have applied this result to develop linear action map teleoperation grasping control schemes \cite{matrone2012realtimemyoelectric,odest2007twoDsubspace,artemiadis2010emgbasecontrol} and planning algorithms \cite{ciocarlie2009handposture}. Linear maps are advantageous because they are easy to analyze and provide intuitive mappings for teleoperation. However, globally linear approaches assume all task-relevant high-DOF commands exist in a single low-dimensional subspace, and need more dimensions for precision movements \cite{artemiadis2010emgbasecontrol}.

\begin{figure}[tb]
    \vspace{0.2in}
    \hspace{-0.1in}
    \centering
    \includegraphics[width=0.35\textwidth]{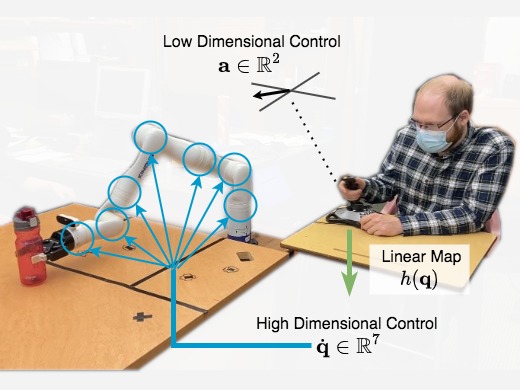}
    \caption{A user controls a 7-DOF robotic manipulator through low dimensional actions of a 2-DOF joystick. The state-conditioned linear mapping $h(\mathbf{q}) \in \mathbf{R}^{7 \times 2}$ transforms the joystick inputs $\mathbf{a}$ to high dimensional motor commands  $\dot{\mathbf{q}}$.}
    \label{fig:overview}
\end{figure}

Other research has investigated nonlinear action mappings as an alternative. These mappings are typically learned with a conditional neural autoencoder (CAE) \cite{losey2021learnlatentLONGpaper,losey2019controlling}. Instead of a single linear map,  low-dimensional actions are input to a state-dependent neural decoder to predict the corresponding high-dimensional motion. Assistive robotic research has demonstrated that CAE models enable higher success rates and faster completion times in assistive eating tasks compared to alternative systems  \cite{jeon2020SharedAW,karamcheti2021learningvisuallyguided,Li2020LearningUM,losey2019controlling,losey2021learnlatentLONGpaper}. However, CAEs are difficult to analyze because of their nonlinear structure. Authors have relied on engineering auxiliary loss terms
to implicitly add desirable user-control properties \cite{Li2020LearningUM,allshire2021laser}, but still require empirical evaluation to verify enforcement on deployment.

Given the potential benefits of both methods and their applications, we are interested in answering the question: \textit{Can we combine the analytic advantages of linear mappings with the flexibility of nonlinear neural networks in a single model?} 
We answer in the affirmative, summarizing our contributions as follows: 1) We propose \textit{State Conditioned Linear Maps} (SCL maps), which use neural networks to predict a local linear basis for control. 2) We formulate the soft reversibility property for action mappings and prove that by design SCL maps guarantee this property. 3) The results of two teleoperation user studies reveal that participants found SCL maps more effective than PCA and CAE baselines and competitive with mode switching.

\section{Background} \label{sec:background}

In this section, we summarize background information and define relevant notations.
Throughout this paper, we use small bold letters for vectors (e.g., $\mathbf{s, a}$)
and capital bold letters for matrices (e.g., $\mathbf{H}$). All vectors are assumed to be columns. Additional notations will be described as they become relevant.

\subsection{Problem Formulation}

We use $\mathbf{o}_{r}$ for robot observation information,  and $\mathbf{o}_{h}$ for agent observations. 
In our experiments, $\mathbf{o}_{r}$ are joint angles $\mathbf{q}$ and gripper state $c\in [0, 1]$, but $\mathbf{o}_{r}$ could also be e.g. end effector location   $\mathbf{x} \in \mathbb{R}^{3}$, or other state. We define $\mathbf{o}_{h}$ as the sensory inputs of an intelligent agent which can include visual, auditory, haptic, and other sensory inputs. 
%
The agent provides low-dimensional control inputs $\textbf{a}$ which get transformed into high-dimensional robotic commands $\dot{\textbf{q}}$ to complete certain manipulation tasks.
We assume that the tasks have the Markov property, such that the action can be determined based purely on the current observation.
The policy $\pi(\mathbf{o}_h) = \mathbf{a}$ denotes the mapping\footnote{Throughout this work $\pi$ is always embodied by a human user.} from agent observation $\mathbf{o}_h$ to the low-dimensional action $\mathbf{a}$.
We are interested in learning state specific mappings $g(\mathbf{o}_r, \textbf{a}) = \dot{\textbf{q}}$. 
We assume that the state transition is deterministic, and define the transition function as $\mathbf{q'} = T(\mathbf{q}, \mathbf{a}; g)$ which outputs the next state (joint angles) $\mathbf{q'}$ when taking a low-level action $\mathbf{a}$ from state $\mathbf{q}$.
The transition function depends on the state-conditioned map $g$ that is trained from demonstration data, and fixed during control.
We write $T(\mathbf{q}, \mathbf{a})$ omitting $g$ when it is clear from context that $g$ is implied. 

\subsection{Previous Action Mapping Functions}

Prior works have modeled $g(\mathbf{o}_r, \mathbf{a})$ with PCA and conditional autoencoders. Principal component analysis reduces the dimensionality of data by calculating the first $d$ principal components and using them as an orthonormal basis with which to represent the data. We denote the matrix with the first $d$ principal components as its columns as $\Sigma \in \mathbb{R}^{m \times d}$ \cite{pearson1901fittingplanesPCA}. 
The action map can then be defined as  $g(\mathbf{q}, \mathbf{a}) = \Sigma \mathbf{a} = \hat{\dot{\mathbf{q}}}$, where the output is independent of $\mathbf{q}$. A conditional autoencoder is an unsupervised learning model that attempts to summarize high-dimensional data $\dot{\mathbf{q}} \in \mathbb{R}^{m}$ in some lower dimensional space, while conditioning the decoder on additional context information $\mathbf{o}_r$. The encoder $f(\mathbf{q}, \dot{\mathbf{q}}) = \textbf{a}$  compresses the data, and a decoder reconstructs the data $g(\mathbf{o}_r, \mathbf{a}) = \hat{\dot{\mathbf{q}}}$. Both $f$ and $g$ are neural networks with parameters  $(\phi, \theta)$ respectively. They are trained with mean squared error loss: $\min_{\phi, \theta} \mathbb{E}[\|\dot{\mathbf{q}} - \hat{\dot{\mathbf{q}}}\|^{2}_{2}]$.  At deployment, the decoder is the action map and $\mathbf{a}$ corresponds to user inputs.

\section{Methods} \label{sec:methods}

This section describes our method for learning to control high DOFs with low-dimensional actions. We assume that there are available training tuples $\{(\textbf{o}_{r}, \dot{\textbf{q}})\}_{i}$ to learn state conditioned action mappings. We outline the model, training procedure, and a post-processing step.  

\subsection{State Conditioned Linear Mappings}

\begin{figure}
 
\vspace{0.2in}
\hspace{0.2in} 
\begin{subfigure}{.45\textwidth}
  \centering
  \includegraphics[trim=0mm 0mm 0mm 0mm, clip=true,width=0.85\linewidth]{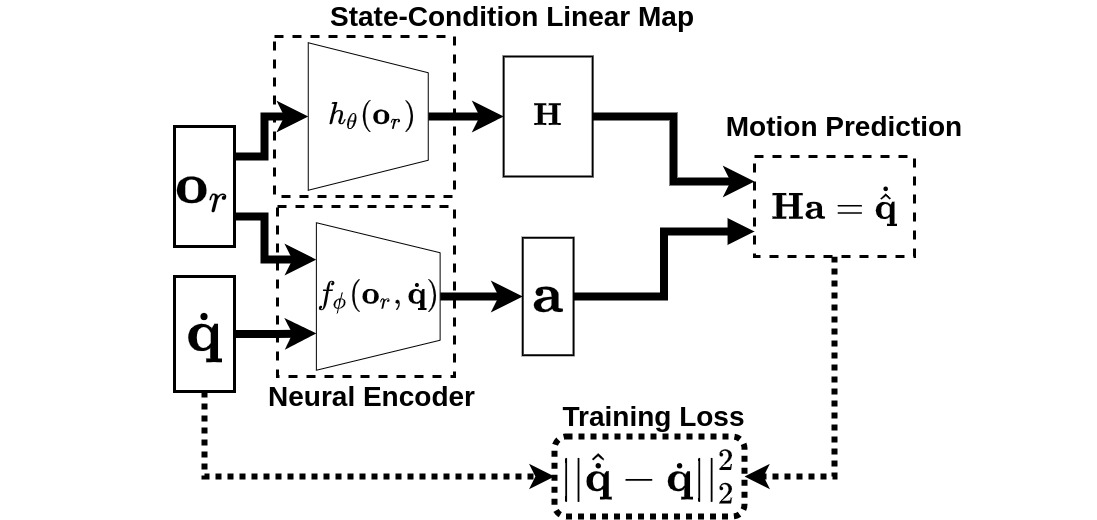}  
  \caption{\textbf{Training}}
\end{subfigure}

\hspace{0.2in} 
\begin{subfigure}{.45\textwidth}
  \centering
  \includegraphics[trim=0mm 0mm 0mm 0mm, clip=true,width=0.9\linewidth]{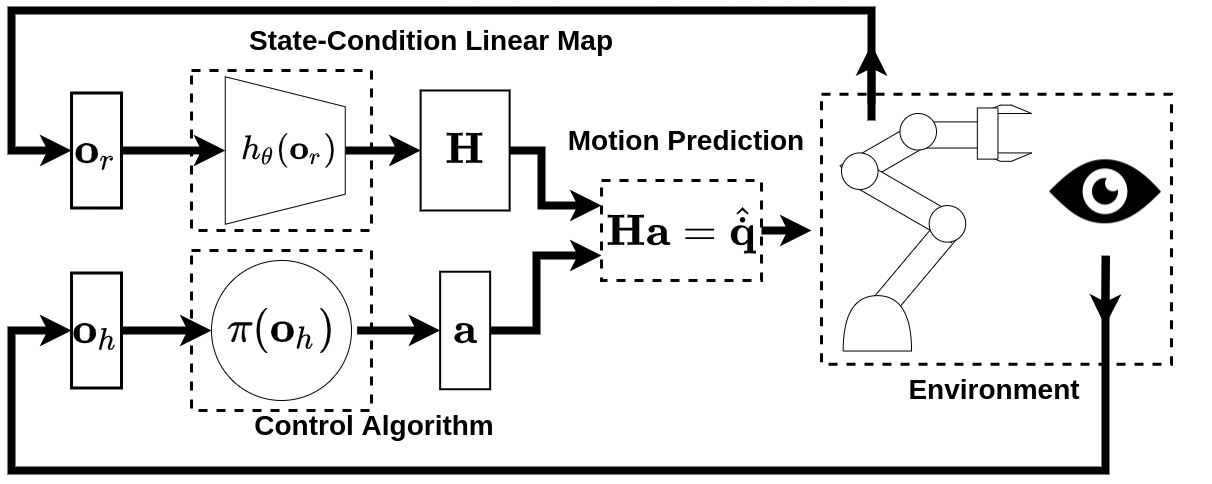} 
  \caption{\textbf{Deployment}}
  \label{fig:sub-second}
\end{subfigure}
\caption{State conditional linear maps are trained as autoencoders. At deployment, the encoder is replaced with the appropriate control policy. In our teleoperation user study, the control policy $\pi(o_h)$ is embodied by a human.}
\label{fig:system}
\end{figure}

A linear action map is desirable because it is easier to mathematically analyze. Knowing how to utilize a robot's context for a task can help determine the appropriate behavior, but this may be a nonlinear function. These two observations lead us to build neural networks ($h: \mathbb{R}^{m} \rightarrow \mathbb{R}^{m \times d};\mathbf{o}_r\mapsto \mathbf{H}$) that output a linear map ($\mathbf{H}:\mathbb{R}^d\rightarrow\mathbb{R}^m$): 
\begin{equation}
\hat{\dot{\textbf{q}}} = \mathbf{H}\mathbf{a} = h_{\theta}(\textbf{o}_{r})\textbf{a}.
\end{equation}
Here, $\hat{\dot{\textbf{q}}} \in \mathbb{R}^{m}$ is the predicted high dimensional robotic velocity command and $\textbf{a} \in \mathbb{R}^{d}$ (a column vector) is a user's low dimensional action. The matrix $\textbf{H} \in \mathbb{R}^{m \times d}$ is the state-conditioned linear map (SCL map). It represents a linear subspace of high-dimensional manipulation commands controllable by low-dimensional actions. State-conditioning allows the model to contextualize high-dimensional manipulation commands specific to the robotic state $\textbf{o}_{r}$ \cite{losey2019controlling}. The user's actions determine a linear combination of the columns of the SCL map to form a desired high dimensional command. Linear map prediction has been previously considered for developing control algorithms \cite{e2c2015watter,zhao2009NNmultifingerJacobian,lyu2020datadrivenLearningofJacobian,lyu2018NNvisionjacobianpred,przystupa2021AnalyzingNJ}.

\subsection{Training SCL Maps}

To train $h_{\theta}(\textbf{o}_{r})$, we optimize the system as a neural autoencoder with mean squared error loss:
\begin{equation}
    \min_{\phi, \theta} ||\dot{\textbf{q}} - h_{\theta}(\textbf{o}_r)f_{\phi}(\textbf{o}_r, \dot{\textbf{q}}) ||^{2}_{2},
\end{equation}
where $\phi, \theta$ are optimized jointly.
The purpose of $f$ is to produce appropriate actions that combine the SCL basis vectors to reconstruct the joint velocities. At deployment $f_{\phi}(\mathbf{o}_{r}, \dot{\mathbf{q}} ) = \textbf{a}$ is discarded and is replaced by an  agent $\mathbf{a} \sim \pi(\mathbf{o}_{h})$. Figure~\ref{fig:system} shows the training and deployment stages of SCL maps.

\subsection{Enabling Diverse Actions}

 A learned SCL map $\textbf{H}$ may be sufficient to reconstruct high DOF control commands during training, but not facilitate exploration of the state space at deployment. 
 Consider a two-joint robot. If $\textbf{H}$ has columns $[1, 0]^{\top}$, and $[1, 0.1]^{\top}$, then the columns span all of $\mathbb{R}^2$. Yet, both $\textbf{a} = [1, 0]^{\top}$ and $\textbf{a} = [0, 1]^{\top}$ will give roughly the same motion, which could make it difficult to reach certain joint configurations.   
We address this issue by applying the Gram-Schmidt process on the columns of $\textbf{H}$ to create an orthonormal representation $\textbf{H}'$ \cite{Schmidt1907gramschmidt}. The columns of $\textbf{H}'$ form an orthonormal basis that spans the same sub-space as $\textbf{H}$, but can lead to more varied actions. Using the mapping $\textbf{H}'$, it is guaranteed that orthogonal inputs map to orthogonal joint velocity commands. 
\section{Properties of SCL}\label{sec:casestudy}

In this section we show theoretically and empirically how SCL maps guarantee both \textit{proportionality} and \textit{reversibility} which are desirable properties in robotic manipulation. We use $\Psi(\textbf{q}) = \textbf{x}$ as the kinematics function of the end-effector's position and orientation \cite{Craig86introrobotics}.
The transition operator is defined as $T(\textbf{q},\textbf{a};g) = \textbf{q} + g(\textbf{q}, \textbf{a}) = \textbf{q} + h(\textbf{q})\textbf{a}$.\footnote{In practice, the inputs of $g$ are not limited to joint angles and can still contain additional task specific information.} We dropped $\theta$ in $h(\textbf{q})$ to ease notations.

This section was motivated by the work of Li et. al. \cite{Li2020LearningUM}, whose work we based our  properties on, but is otherwise different for several reasons. Their method is semi-supervised targeted for a human operator and requires them to label subsets of transition data by providing the desired joystick input. Our method is unsupervised thus eliminating the labeling process. Their system also needs additional loss terms to enforce these properties, which require additional hyperparameter tuning.



\paragraph{Proportionality} For scalar $\alpha \in \mathbb{R}$ the resulting action $\textbf{a}' = \alpha \textbf{a}$ will lead to a proportional change in the end effectors current position: 
\begin{equation*}
    \alpha \| \Psi(T(\textbf{q}, \textbf{a}; g)) - \Psi(\textbf{q})\| \approx \|\Psi(T(\textbf{q},\alpha \textbf{a}; g)) - \Psi(\textbf{q})\|.
\end{equation*}
Intuitively, the end-effector is expected to move in proportion to the magnitude of the input. As SCL maps predict a linear weight matrix, proportionality is locally achieved (where kinematics $\Psi$ is approximately linear) by design. Suppose we have that $\textbf{a}' = \alpha \textbf{a}$ for $\alpha \in \mathbb{R}^+$. This will lead to a proportional increase of the response ($\textbf{H}\textbf{a}' = \alpha \textbf{H}\textbf{a} = \alpha \dot{\textbf{q}}$). 

\begin{figure}[t]
    \centering
    
    \vspace{0.25in}
    \hspace{0.1in}
    \includegraphics[width=0.4\textwidth]{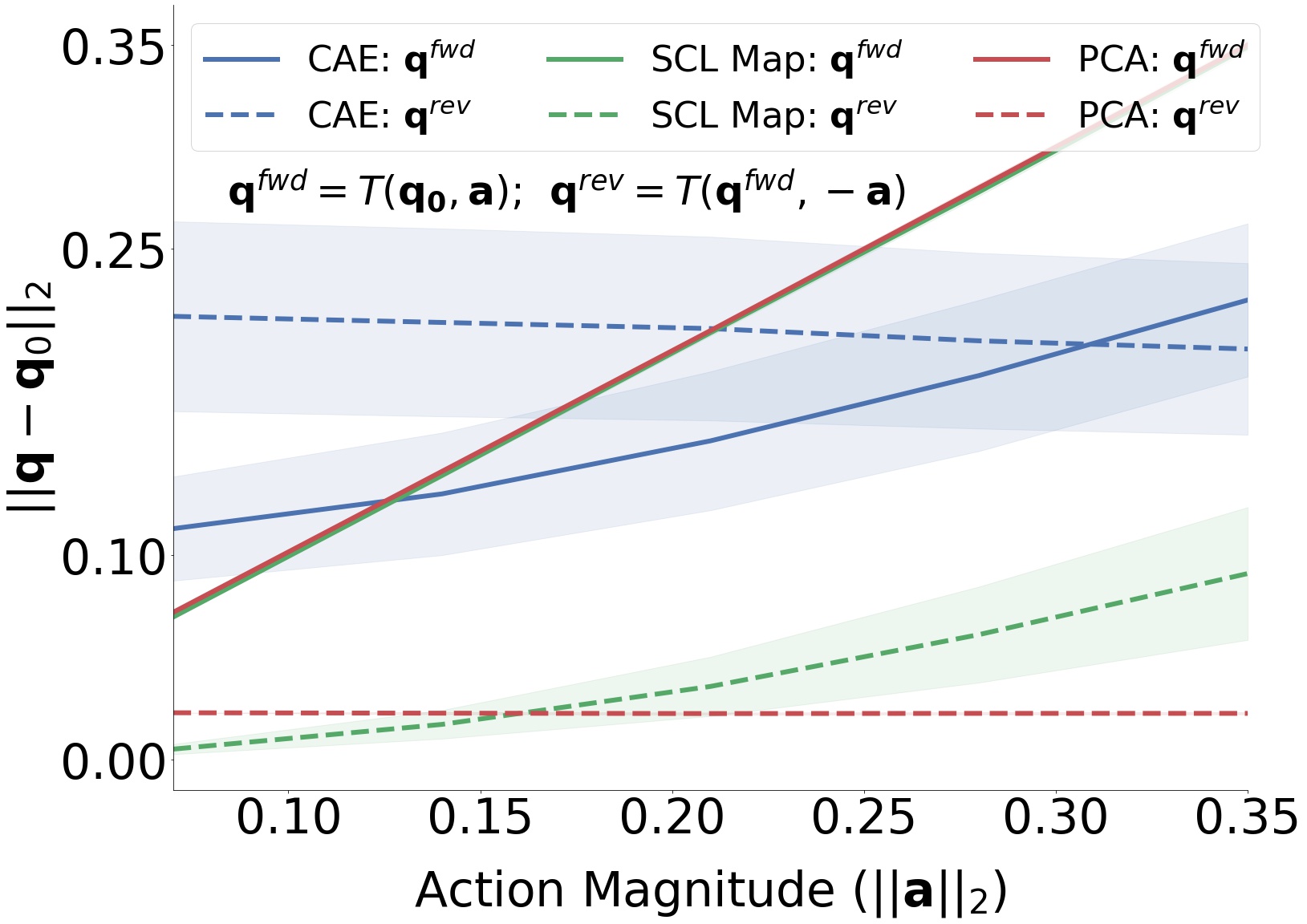}
    \caption{Empirical results from an experiment demonstrating proportionality and soft reversibility of SCL maps on the Kinova Gen-3 lite robot. The line and shaded area show the mean and one-half standard deviation from 100 runs (10 trained models, 10 random states), respectively. Note $||\mathbf{q}^{fwd}-\mathbf{q}_0||_2$ for SCL and PCA overlap.
    }
    \label{fig:softreversibility}
\end{figure}
\paragraph{Soft Reversibility}

For two states $\textbf{q}_{i}$ and $\textbf{q}_{j}$ and an action $\mathbf{a}\neq 0$ such that $\textbf{q}_{j} = T(\textbf{q}_{i}, \textbf{a} ; g)$, ``Reversibility'' means that:
\begin{equation}
    \textbf{q}_{j} = T(\textbf{q}_{i}, \textbf{a} ; g) \Rightarrow \textbf{q}_{i} = T(\textbf{q}_{j}, -\textbf{a}; g). \nonumber
\end{equation}

For this definition to hold for SCL maps, $h(\textbf{q}_j)$ must be equivalent to $h(\textbf{q}_i)$, which unfortunately is not guaranteed.
Instead, we show that SCL satisfies a ``soft reversibility'' property: the state $\textbf{q}_{k}$ reached by the inverse action from $\textbf{q}_{j}$ will be closer than $\textbf{q}_{j}$ to $\textbf{q}_{i}$.

\begin{theorem}[Soft Reversibility]
Let $\|\textbf{a}\|_2 < 1$ and $T(\textbf{q}, \textbf{a}; g) = \textbf{q} + g(\textbf{q}, \textbf{a}) = \textbf{q} + h(\mathbf{q}) \mathbf{a}$. 
$h(\mathbf{q}) \in\mathbb{R}^{m \times d}$ is a matrix  transformed from the vector form of hidden layer activations $v(\mathbf{q}) \in \mathbb{R}^{d m}$.
Suppose $v$ is Lipschitz continuous for some $L_{v} \leq 1$, that is, $\| v(\textbf{q}_{j}) - v(\textbf{q}_{i}) \|_2  \leq L_v \| \textbf{q}_{j} - \textbf{q}_{i} \|_2$.

Then for some $\textbf{q}_{j} = T(\textbf{q}_{i}, \textbf{a}; g)$ we have:
\begin{equation*}
    \|T( \textbf{q}_{j}, -\textbf{a}; g) - \textbf{q}_{i}\|_2 
    < \|\textbf{q}_{j} - \textbf{q}_{i}\|_2
\end{equation*}
\label{thm}
\end{theorem}

\begin{figure}[h]
\vspace{0.1in}

  \centering
  \includegraphics[width=0.7\linewidth]{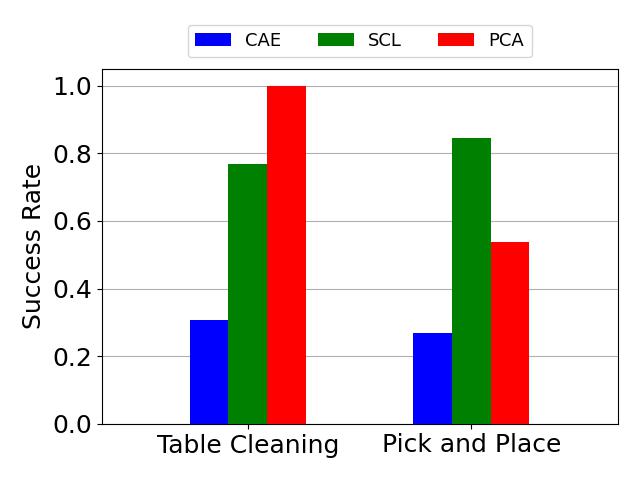}  
  \label{fig:tablecleanlikert}
\caption{The number of trials in which participants successfully completed the Table Cleaning and Pick and Place tasks.}
\label{fig:usertrial_results}
\end{figure}
\begin{proof}
From our definition of $T(\textbf{q}, \textbf{a}; g)$:
\begin{align}
\textbf{q}_{j} & = \textbf{q}_{i}+h(\textbf{q}_{i})\textbf{a}_i  \nonumber\\ 
\textbf{q}_{k} & = \textbf{q}_{j} + h(\textbf{q}_{j})\textbf{a}_{j} \label{eq:snn}
\end{align}
where $\textbf{q}_{i}, \textbf{q}_{j}, \textbf{q}_{k}$ are three consecutive states,
$\textbf{a}_i$ and $\textbf{a}_{j}$ are the actions at current time step and the next time step respectively.
If the action $\textbf{a}_{j}$ is the reverse action $\textbf{a}_{j} = -\textbf{a}_i$, we can rewrite Eq. \eqref{eq:snn} as
\begin{align*}
\textbf{q}_{k} & = \textbf{q}_{i} + h(\textbf{q}_{i})\textbf{a}_i + h(\textbf{q}_{j}) (-\textbf{a}_i) \\
& = \textbf{q}_{i} + (h(\textbf{q}_{i})-h(\textbf{q}_{j}))\textbf{a}_i 
\end{align*}
From the Lipschitz continuity assumption, it follows  that:
\begin{align}
\nonumber
\| \textbf{q}_{k} - \textbf{q}_{i} \|_2 & = \| (h(\textbf{q}_{i}) -h(\textbf{q}_{j})) \textbf{a}_i  \|_2 \\ 
& \leq \| h(\textbf{q}_{i}) - h(\textbf{q}_{j}) \|_2 \: \| \textbf{a}_i \|_2 \\ 
& \leq \| h(\textbf{q}_{i}) - h(\textbf{q}_{j}) \|_F \: \| \textbf{a}_i \|_2 \\ 
& = \| v(\textbf{q}_{i}) - v(\textbf{q}_{j}) \|_2 \: \| \textbf{a}_i \|_2 \\ 
 & \leq  L_v  \| \textbf{q}_{i} -\textbf{q}_{j}\|_2 \: \| \textbf{a}_i  \|_2 \\
 & \leq  (L_v \| \textbf{a}_i  \|_2 ) \: \| \textbf{q}_{i} -\textbf{q}_{j}\|_2 \\
 & <  \| \textbf{q}_{i} -\textbf{q}_{j}\|_2
 \label{eq:dss}
\end{align}
for any pairs of joint angles in a robot's set of joint configurations $\textbf{q}_{i}, \textbf{q}_{j} \in \mathcal{Q}$ such that $\textbf{q}_{j} = T(\textbf{q}_{i}, \textbf{a}_i ; g)$, where $\mathcal{Q}$ is the set of joint angles of the robot.
We can tell from Eq. \eqref{eq:dss} that if $L_v \| \textbf{a}_i  \|_2 < 1 $, taking the reverse action $\textbf{a}_j=-\textbf{a}_i$ brings the robot to a state closer to $\textbf{q}_i$ than before. 

\end{proof}




\paragraph{Experiments Validating Properties}
We empirically validate our theoretical results for SCL maps properties on a \textit{Kinova Gen-3 lite} \cite{kinovakortex}, comparing the properties of SCL maps to those of CAE and PCA models. For the SCL and CAE algorithms, we trained 10 models each using the Adam optimizer for 200 epochs, with a learning rate of $1e\text{-}3$ and mini-batches of 256. 
Both SCL and CAE were trained with encoders that had two hidden layers each, with 256 neurons and tanh activation functions. The decoders had similar configurations, except that the SCL decoder outputs a matrix as described in Section~\ref{sec:methods}. During training of the SCL maps, we enforced Lipschitz continuity with $L_v=1$ by applying the Lipschitz training procedure in Algorithm 1 of Gouk et. al. \cite{gouk_regularisation_2021}. We forced the linear maps $\mathbf{H}=h_\theta(\textbf{q})$ to be orthonormal with the Gram-Schmidt process during training and deployment. 

To compare the proportionality and reversibility properties of each algorithm, we generate an angle $\theta \sim \mathcal{U}(0,2\pi)$ which we transform into actions $\textbf{a} = \alpha [\cos(\theta), \sin(\theta)]^\top$. We control the norm of the actions with the scalar $\alpha$. For each value of $\alpha$, we first apply the action $\textbf{a}$ for one second, stop the motion, and then apply the inverse action $-\textbf{a}$. We denote $\textbf{q}^{fwd}$ and $\textbf{q}^{rev}$ as the joint configurations of the robot after the forward and inverse actions, respectively. For each run of the experiment we measure first the distance from the start state $\textbf{q}^{0}$ to $\textbf{q}^{fwd}$, and then the distance of $\textbf{q}^{rev}$ to $\textbf{q}^{0}$.


Our results are shown in Fig.~\ref{fig:softreversibility}. Ten random states from our demonstration data were chosen as initial states, and each algorithm (CAE, SCL, PCA) was used to compute joint velocities given a random action with increasing magnitude.
Although we do not consider the Gram-Schmidt process in our proof for Theorem \ref{thm} (we leave this to future work), the results of this experiment on the Kinova manipulator serve as empirical evidence that soft reversibility does hold in practice for SCL maps transformed by the Gram-Schmidt process.

\begin{figure}
\vspace*{1.0cm}
    \centering
    \hspace*{-1cm}
    \vspace*{-0.2cm}
    \includegraphics[trim=5mm 1mm 2mm 2mm, clip=true, width=0.45\textwidth]{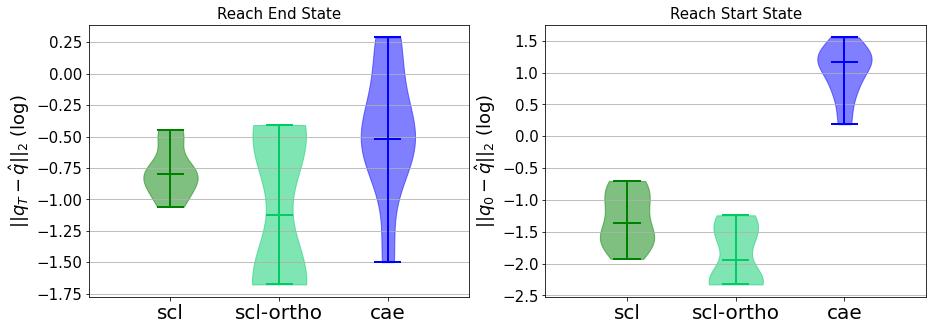}
    \caption{Violin plots of euclidean distances across human prior loss combinations for training. Experiments were on simulated 5-DOF planar reaching task. Reported in log scale for visualization.} 
    \label{fig:reachingmetrics}
\end{figure}

\begingroup
\begin{figure*}[htp] 
\vspace{0.5in}
\hspace{0.00in}
    \begin{minipage}[t]{0.33\textwidth}
        \centering
        
        \includegraphics[trim=0mm 0mm 5mm 5mm, clip=true, width=\textwidth]{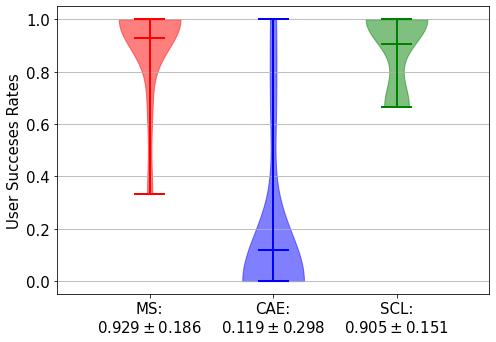}
        \caption{Percentages User Completions. }
        \label{fig:pouringcompletion_rates}
    \end{minipage}
    \begin{minipage}[t]{0.33\textwidth}
        \centering
        
        \includegraphics[trim=1mm 1mm 1mm 1mm, clip=true, width=0.85\textwidth]{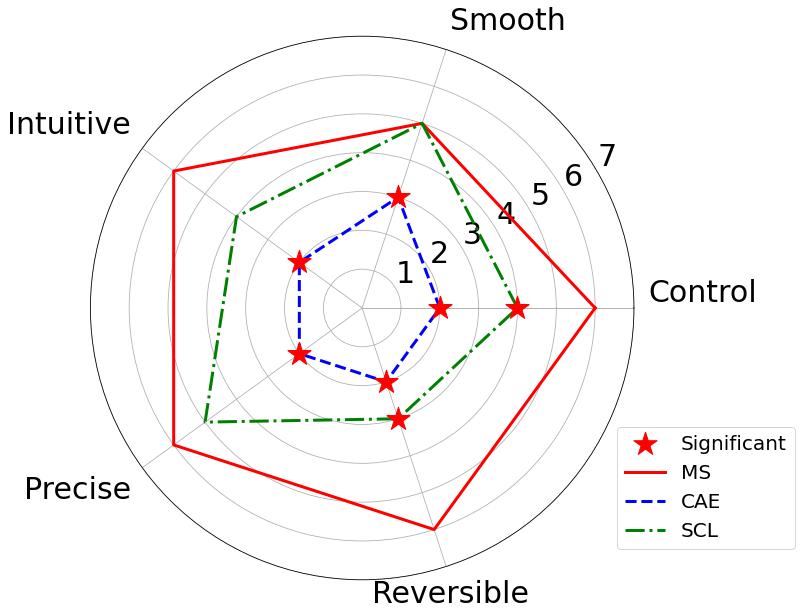} 
        \caption{Median Likert Scale results.  } 
        \label{fig:likertpourresult}
    \end{minipage}
    \begin{minipage}[t]{0.33\textwidth}
    \centering
        \includegraphics[trim=2mm 2mm 1mm 1mm, clip=true, width=0.9\textwidth]{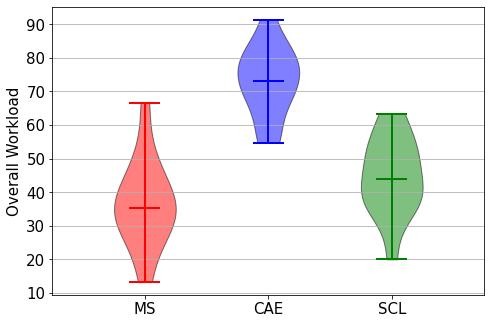}
        \caption{NASA Workload Index.  }
        \label{fig:nasaworkload}
    \end{minipage}
    
\caption*{\textbf{Pouring User Study Results.} Figure~\ref{fig:pouringcompletion_rates} Violin plots of completion with mean. Figure~\ref{fig:likertpourresult} Radar plot with stars indicating significance wrt mode switching. Figure~\ref{fig:nasaworkload} Violoin plots of NASA TLX scores with median.}
\label{fig:pouringexperiments}
\end{figure*}
\endgroup
\section{User Study} \label{userstudy}
We conduct two sets of user studies to validate the efficacy of SCL as a learnable action mapping approach. We chose to evaluate SCL in teleoperation because of previous research in assistive robotics \cite{losey2019controlling, losey2021learnlatentLONGpaper}.  The first user study compares across the spectrum of action mapping approaches on two synthetic tasks that empirically could be solved with actions existing in a two dimensional subspace. The second user study pushes the limits of SCL when compared to an assitive robotic's mode switching interface as well as a more advanced CAE model. The second user study task could not be solved with the first two principal components of PCA and was excluded. All experiments were conducted on a Kinova Gen3 lite robot with a control rate of 40 Hz \cite{kinovakortex}. Reported statistical significance used a 0.05 p-value. A Kruskal-Wallis test was performed before reported Dunn test significance. 

\begin{figure}
    \centering
    \begin{tabular}{@{}cc}
         {\includegraphics[height=2.35cm, trim={3cm 20cm 0 10cm},clip]{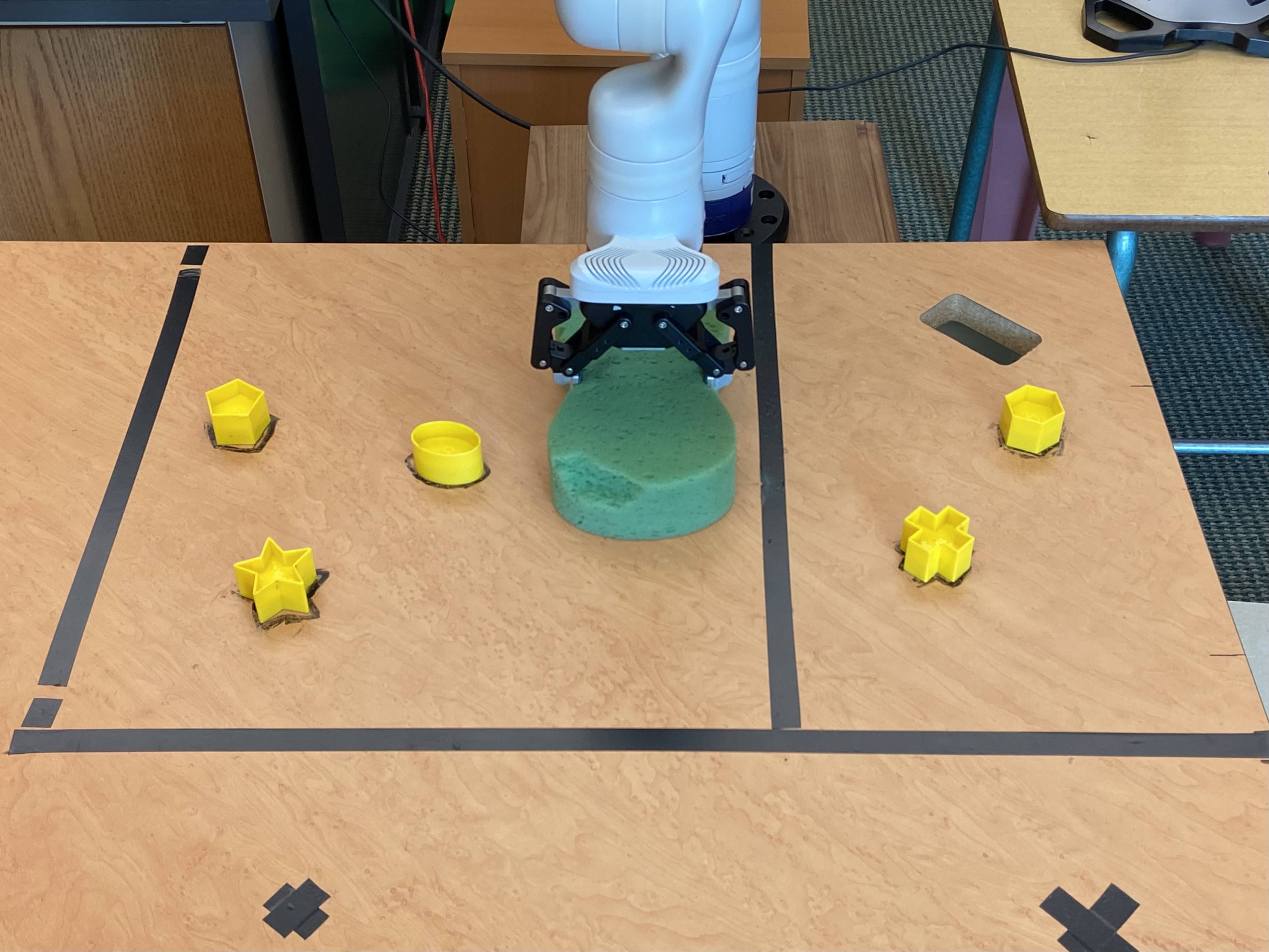}}&
         {\includegraphics[height=2.35cm,trim={20cm 2cm 2cm 5cm},clip=True]{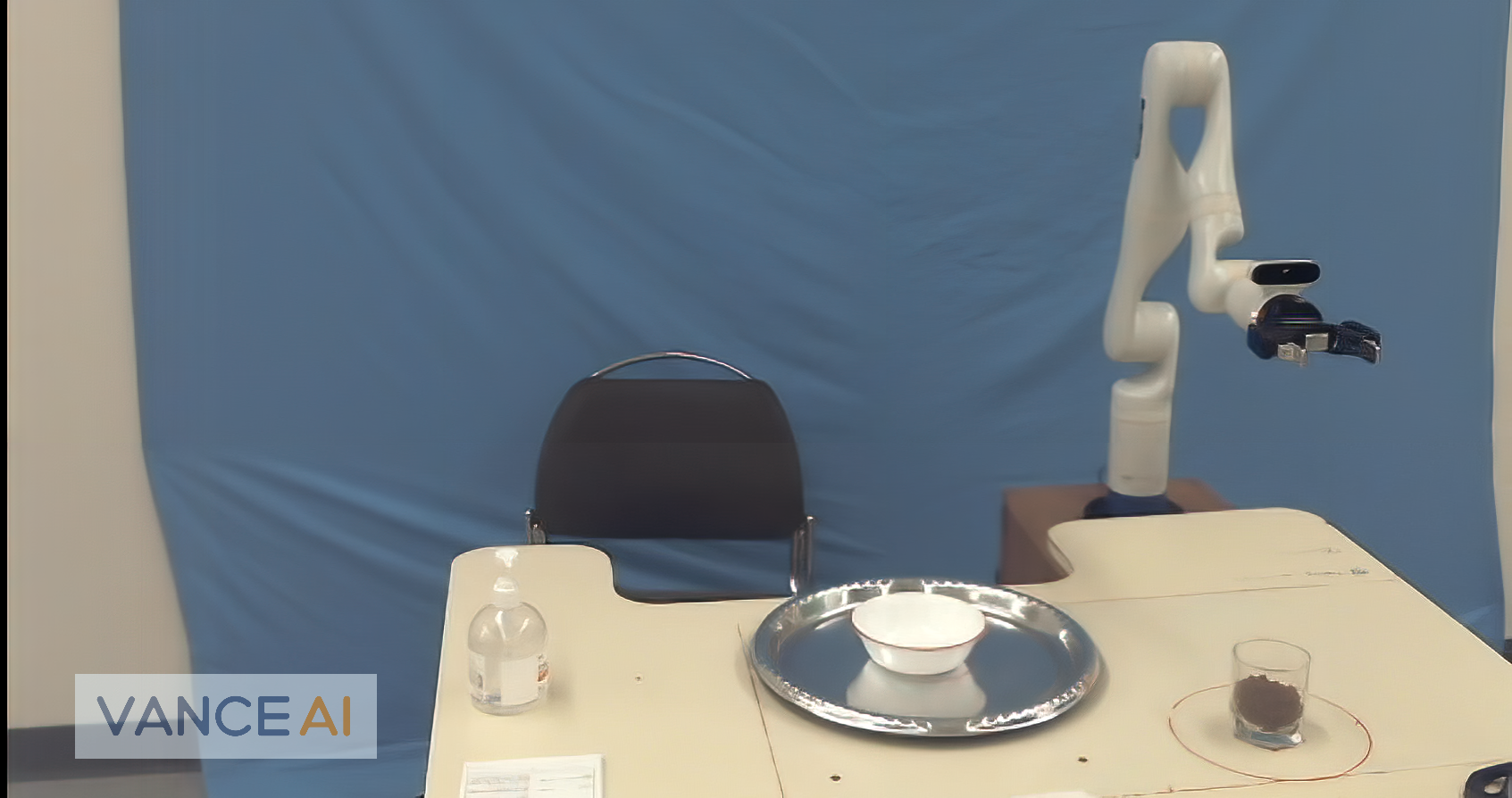}}
     \end{tabular}
    \caption{Table Cleaning and Pouring Experiments. 
    }
    \label{fig:tablecleaningtask}
    \vspace{-0.5cm}
\end{figure}

\subsection{Latent Action User Study}
Our first user study compared SCL against other action mapping methods in the literature. We chose tasks that can be solved with 2D and 3D Cartesian control, which we refer to as \textit{Table Cleaning} and \textit{Pick and Place} in this section. Despite being simple, these tasks allow us to empirically evaluate the efficacy of SCL in general compared to existing action mapping approaches.

\paragraph{Experimental Set-up}
The Table-Cleaning task set up is shown in Figure \ref{fig:tablecleaningtask}. Participants were asked to push five small objects on a table out of a designated boundary in 60 seconds. For the table cleaning task, we collected 5 demonstrations of circular scrubbing motions on a table. Three trajectories were used to train the PCA, SCL and CAE models, and the remaining two were held out for validation. The total amount of training data was 509 ($\textbf{q}$,$\dot{\textbf{q}}$) tuples.

 The set up for the Pick and Place trials can be seen in Figure \ref{fig:overview}. The task was to reach from a home configuration to grasp the bottle, which was initially placed on the black ``x" marked on the table within 90 seconds. The participants were then instructed to move the bottle to another fixed, marked position on the table. Grasping was controlled by a trigger on the joystick controller. We collected eight demonstration trajectories for the pick and place task. We used six trajectories for training and two for validation of each model. The six training trajectories accounted for 2001 ($\q$,$\dot{\q}$) tuples. 
 For each task, participants were given five minutes to practice, and two attempts at the task. We had 13 participants (4 females and 9 males, age 23 - 33). We followed a single-blind experimental set-up, randomizing the order of models each participant used for each task.\footnote{The studies were approved by University of Alberta Research Ethics and Managements (Pro00054665).}


\paragraph{Models}
We compared SCL against PCA \cite{odest2007twoDsubspace,artemiadis2010emgbasecontrol} and a conditional autoencoder (CAE) \cite{losey2021learnlatentLONGpaper,losey2019controlling}. 
Respectively, they represented a linear and non-linear approach for learning low-to-high dimensional mappings for control. Our CAE model is based on the open-source version of Karamcheti et. al. \cite{karamcheti2021learningvisuallyguided}. The conditional autoencoder and SCL maps were trained on demonstration data to fit the mean-squared error between the recorded and reconstructed joint velocities (i.e. with unsupervised learning). The SCL and CAE models were trained with the same data, learning parameters, and architectures described in Section \ref{sec:casestudy}.c. For each algorithm we chose the best model (out of ten) based on a validation loss for deployment. Participants used a 2-DOF joystick to provide low-dimensional control commands with inputs mapped as actions $\textbf{a}=[a_0, a_1]^\top$, with $\|\mathbf{a}\| \leq 1$. 

\paragraph{Results and Discussion}

Our empirical results are included in Figure~\ref{fig:usertrial_results} where we report success rates. Our results suggest that PCA worked well for participants in the table-cleaning task, but it failed to perform consistently in the Pick and Place task. There was a qualitative difference in the motions required to reach the bottle versus the motions needed to move the bottle to its destination (i.e. different motions were needed to ``pick" vs. to ``place"). PCA was forced to capture all variation in joint velocities for both pick and place motions with just two principal components. SCL, on the other hand, was free to adapt the subspace of joint velocities, given the current configuration of the manipulator. This meant that the joint velocity commands available to the participants (via output of action map $g(\mathbf{o}_r,\mathbf{a})$) could be entirely different when reaching for the bottle, versus after the bottle had been grasped. We observed that the ability of SCL to use the current state to adapt the action maps for locally useful actions resulted in a statistically significant advantage in the pick and place task, in terms of the number of successful trials for each participant (Fig. \ref{fig:usertrial_results}).

Based on our observations, the low success rate of the CAE model could be attributed to two primary reasons. First, the CAE internally uses affine mappings as opposed to linear mappings. When a user's input is $\textbf{0}\in\mathbb{R}^2$ (no input), the joint velocity generated will be determined by the bias terms in the hidden layers of the model. The SCL method does not suffer from this problem, because it instead transforms actions with a linear map.

 Second, we observed that many CAE trials resulted in failure if users navigated to joint configurations outside of the training trajectory distribution. In these states, the robot behavior became unpredictable, and in many cases counter-productive (e.g. all actions appeared to move the end-effector away from the bottle). These issue of unpredicted behavior have been previously seen in the literature \cite{losey2019controlling}. In contrast,  the soft reversibility property of SCL enabled participants to recover from configurations outside of the training distribution.

\subsection{Simulation Experiments for Human Prior Configuration}
Prior to our second user study, we sought to improve the performance of CA by conducting simulations in which we experimented with different training loss formulations. We re-implemented the \textit{consistency}, \textit{proportionality}, and \textit{reversibility} loss functions proposed in Section 4.B of Li et al \cite{Li2020LearningUM}. We performed a grid search over combinations of these loss functions as well as the sampling range for the \textit{Proportionality} hyperparameter $\alpha \in [0.1, 1.0]$ and chose a fixed temperature parameter of $\gamma = 2.0$ in the \textit{Consistency} loss. The reported experiments used a 5 DOF planar reaching task where targets varied along a single line. Trajectories were collected with inverse Jacobian control. For each configuration, we trained 10 models with the same hyperparameters as our first user study (see section section \ref{sec:casestudy}.c), over 1000 epochs. We report results from the point in training with the lowest validation loss for each configuration. We repeated these experiments for SCL with and without Gram-Schmidt orthonormalization.

We evaluated each system by simulating a greedy user that chose the best action (in the following sense) to reach a target joint configuration. Actions were determined by generating 4096 random actions from an Isotropic Gaussian and picking the action that minimized the following loss:
\begin{equation}
    \textbf{a}^{*} = \min_{\textbf{a} \sim \mathcal{N}(0, I)} || \textbf{q}^{*} - (\textbf{q}_t + g_{\theta}(\textbf{o}_r, \textbf{a})) ||_{2},
\end{equation}
Where $\textbf{q}^*$ is the goal joint configuration, and we treat the output of the decoder $g_{\theta}$ as joint velocities.
We report the distribution of average errors with respect to the end $\textbf{q}_T$ and returning to the start states $\textbf{q}_0$ on 100 test trajectories for each of the 10 evaluations per hyperparameter configuration in Figure~\ref{fig:reachingmetrics}. We found statistical significance between CAE results and both SCL versions. We found performance comparable to SCL for some configurations of CAE to reach the end state, but results suggest that even with these additional loss terms, SCL is better at reversing actions, as our theory suggested. We repeated this procedure with the collected pouring trajectories, finding the proportionality with $\alpha = 1.0$ and reversibility loss functions to work best and used this configuration for the following user study.
\subsection{Assitive Robotic User Study}

In our second user study, we were interested in comparing SCL to systems that have previously been used for assitive robotics including CAE and a mode switching system (MS) \cite{newman2022harmonic}. We set-up a \textit{Pouring} task (Figure~\ref{fig:tablecleaningtask}), where participants had to pick-up a cup full of beans, pour them into a bowl, and then replace the cup in a designated location on the table within two minutes. This task was highly nonlinear, consisting of several motions that involved both translation and rotation of the robot's end-effector. We collected 10 demonstration trajectories, where 8 were used for training (5200 training tuples) and 2 for validation (1191 validation tuples).

This study included 14 participants (ages 22 - 51, 2 female). Participants were given four two-minute practice trials with each of the control systems to familiarize themelves with the action mappings. We then collected results for each method over three test trials.

We report the average success rates of participants in Figure~\ref{fig:pouringcompletion_rates}. 
We found a significant difference between mean success rates when comparing CAE to both SCL and MS, but not comparing SCL to MS. 
Generally we found users could pick up the glass with any interface, but struggled to pour with the CAE model.  

In addition, we also collected user subjective opinions which included a 5 question Likert scale survey (1 low and 7 high) featured in Figure~\ref{fig:likertpourresult}. We found that only Control and Reversibility were statistically significant by Dunn's Test. We also collected the  NASA Workload Index \cite{hart1988development} featured in Figure ~\ref{fig:nasaworkload}. Again, by Dunn's test we found statistical significance between CAE and both SCL and MS.

Although our results show the promise of SCL in assistive robotic settings, further work is necessary to improve the user experience. With a single mode, SCL performed comparably to mode switching, which exposed three control modes (x-y, z-yaw, and roll-pitch, following \cite{newman2022harmonic}). On average, users switched modes $17.55 \pm 3.91$ times while using the mode switching interface in the pouring task. We conjecture that on even more complex tasks (e.g. dynamic tasks, or tasks requiring simultaneous orientation and position control), the benefits of SCL would be more pronounced over typical assistive robotic systems. In previous works, tasks have often consisted of several sub-tasks and were achievable with CAE models. Interestingly, our results would seem to disagree with existing work on CAE models for shared autonomy. One explanation could be that previous research included additional heuristics to account for CAE limitations. As discussed in several instances, we have found the CAE models move without user input. It's possible to address this heuristically (e.g. send 0 velocity if the action norm is small), but our focus was on achieving properties of the interface end-to-end.

\section{Conclusion}  \label{sec:conclusion}

In this paper, we demonstrated the potential SCL maps have as an approach to learn low-dimensional action mappings. SCL has properties that lie between those of strictly linear and black box non-linear models. We proved a result that SCL maps can provide a form of soft reversibility, where following an action with its inverse is guaranteed to bring the robot closer to its initial joint configuration. In addition, the local linearity of SCL maps guarantees proportional outputs.

Although our teleoperation results show the promise of SCL compared to alternative latent action models, future research is necessary to adapt SCL to specific domains. We feel that SCL has the potential to simplify learning for other robotic control solutions such as reinforcement learning. We also believe that beyond SCL, understanding action manifolds is a promising direction given PCA's effectiveness in our first user study. 
Our results offer promising directions for future action mapping research for robotic manipulation.

\bibliography{references}

\addtolength{\textheight}{-12cm}   

\end{document}